\documentclass[lettersize,journal,twoside]{IEEEtran}
\usepackage{amsmath,amsfonts}

\usepackage[utf8]{inputenc} 
\usepackage[T1]{fontenc}    
\usepackage{url}            
\usepackage{booktabs}       
\usepackage{nicefrac}       
\usepackage{microtype}      
\usepackage{graphics,graphicx,color}

\usepackage{algorithm}
\usepackage{algorithmic}
\usepackage{enumitem}

\usepackage{amsfonts}       
\usepackage{amsmath}       
\usepackage{amssymb}

\usepackage{etoolbox}

\newcommand{\Figref}[1]{Figure~\ref{#1}}  
\newcommand{\figref}[1]{Fig.~\ref{#1}}    

\newcommand{\tabref}[1]{Table~\ref{#1}}

\newcommand{\secref}[1]{Sec.~\ref{#1}} 

\usepackage{xspace}
\makeatletter
\DeclareRobustCommand\onedot{\futurelet\@let@token\@onedot}
\def\@onedot{\ifx\@let@token.\else.\null\fi\xspace}
\makeatother

\usepackage{xr-hyper}
\makeatletter
\newcommand*{\addFileDependency}[1]{
  \typeout{(#1)}
  \@addtofilelist{#1}
  \IfFileExists{#1}{}{\typeout{No file #1.}}
}
\makeatother

\usepackage{xcolor}

\definecolor{ourorange}{HTML}{e19c24}
\definecolor{ourgreen}{HTML}{97b567}
\definecolor{ourred}{HTML}{ec6235}
\definecolor{ourblue}{HTML}{5e81b5}
\definecolor{ourgrey}{HTML}{919191}

\DeclareMathAlphabet{\mathsfit}{\encodingdefault}{\sfdefault}{m}{sl}
\SetMathAlphabet{\mathsfit}{bold}{\encodingdefault}{\sfdefault}{bx}{n}

\newcommand{\norm}[1]{\left\lVert#1\right\rVert}

\usepackage{algorithmic}
\usepackage{array}
\usepackage[caption=false,font=normalsize,labelfont=sf,textfont=sf]{subfig}
\usepackage{textcomp}
\usepackage{stfloats}
\usepackage{url}
\usepackage{verbatim}
\usepackage{graphicx}
\usepackage{cite}
\usepackage[bookmarks=true]{hyperref}
\usepackage{relsize}
\usepackage{multirow}
\usepackage[numbers,sort&compress]{natbib}

\begin{document}

\title{Learning Whole-Body Humanoid Locomotion via \\ Motion Generation and Motion Tracking}
\author{Zewei Zhang$^{1, 4}$, Kehan Wen$^{1}$, Michael Xu$^{2}$, Junzhe He$^{1}$, Chenhao Li$^{1, 3}$, Takahiro Miki$^{1}$, \\ Clemens Schwarke$^{1, 5}$, Chong Zhang$^{1, 3}$, Xue Bin Peng$^{2, 5}$, and Marco Hutter$^{1}$

\thanks{Manuscript received: April 9, 2026; Accepted: June 18, 2026. This paper was recommended for publication by Editor Olivier Stasse upon evaluation of the Associate Editor and Reviewers comments. }
\thanks{$^{1}$Robotic Systems Lab, ETH Zurich {\ttfamily\small \{zewzhang, kehwen, junzhe, chenhli, tamiki, cschwarke, chozhang, mahutter\}@ethz.ch}} 
\thanks{$^{2}$Simon Fraser University {\ttfamily\small \{mxa23, xbpeng\}@sfu.ca}} 
\thanks{$^{3}$ETH AI Center. $^{4}$EPFL. $^{5}$NVIDIA.} 
\thanks{Digital Object Identifier (DOI): see top of this page.}
}

\markboth
{IEEE ROBOTICS AND AUTOMATION LETTERS. PREPRINT VERSION. ACCEPTED JUNE, 2026}
{Zhang \MakeLowercase{\textit{et al.}}: Learning Whole-Body Humanoid Locomotion via Motion Generation and Motion Tracking}

\maketitle

\begin{abstract}
Whole-body humanoid locomotion is challenging due to high-dimensional control, morphological instability, and the need for real-time adaptation to various terrains using onboard perception.
Directly applying reinforcement learning (RL) with reward shaping to humanoid locomotion often leads to lower-body-dominated behaviors, whereas imitation-based RL can learn more coordinated whole-body skills but is typically limited to replaying reference motions without a mechanism to adapt them online from perception for terrain-aware locomotion.
To address this gap, we propose a whole-body humanoid locomotion framework that combines skills learned from reference motions with terrain-aware adaptation.
We first train a diffusion model on retargeted human motions for real-time prediction of terrain-aware reference motions.
Concurrently, we train a whole-body reference tracker with RL using this motion data.
To improve robustness under imperfectly generated references, we further fine-tune the tracker with a frozen motion generator in a closed-loop setting. 
The resulting system supports directional goal-reaching control with terrain-aware whole-body adaptation, and can be deployed on a Unitree G1 humanoid robot with onboard perception and computation. 
The hardware experiments demonstrate successful traversal over boxes, hurdles, stairs, and mixed terrain combinations.
Quantitative results further show the benefits of incorporating online motion generation and fine-tuning the motion tracker for improved generalization and robustness.
\end{abstract}

\section{Introduction} \label{sec:intro}
Developing whole-body perceptive humanoid locomotion remains a challenging yet important step toward expanding the operational range of humanoid robots.
While deep reinforcement learning (RL) has achieved remarkable success in enabling highly dynamic behaviors on quadrupedal systems~\citep{Lee2020, wildanymal, AME, parkourinthewild, anymalmotionprior, AME2}, extending these capabilities to humanoid robots remains considerably difficult, as humanoids possess higher degrees of freedom together with more complex morphological constraints. 
As a result, training an RL-based whole-body humanoid controller solely through reward shaping can be non-trivial and often suffers from inefficient exploration. 
Without explicit structural guidance, vanilla RL-based humanoid controllers frequently converge to locomotion strategies with limited whole-body coordination, making it difficult to acquire coordinated behaviors for challenging terrain traversal.
This limitation becomes especially evident in tasks such as humanoid parkour, where coordinated whole-body movements are essential for traversing large and complex obstacles.

To alleviate the difficulty of reward shaping and inefficient exploration, motion imitation via RL has emerged as a promising alternative~\cite{deepMimic, GMT, beyondmimic, exbody2, twist, omniretarget}. 
By leveraging reference data, motion tracking can efficiently transfer highly coordinated whole-body skills to the robot.
However, pure motion trackers are inherently limited to replaying choreographed trajectories and generally lack the adaptability and reactivity required for operation in unstructured and diverse environments. 
Achieving human-level perceptive locomotion therefore requires a higher-level composition mechanism that can dynamically adjust locomotion style across varied obstacles based on real-time perceptive inputs.

One direct solution for skill composition is to distill multiple specialized teacher policies~\citep{APEX}, or to compose multiple skills from separate reference trajectories within a single imitation-based teacher framework~\citep{PHP,Xloco}. 
By learning distinct expert skills for different terrain conditions, a unified distilled policy can learn to transition among them using exteroceptive observations. 
However, current methods have not yet demonstrated strong scalability to a diverse set of skills or robust generalization to new terrains.

\begin{figure}[t!]
    \centering
    \includegraphics[width=1.0\linewidth]{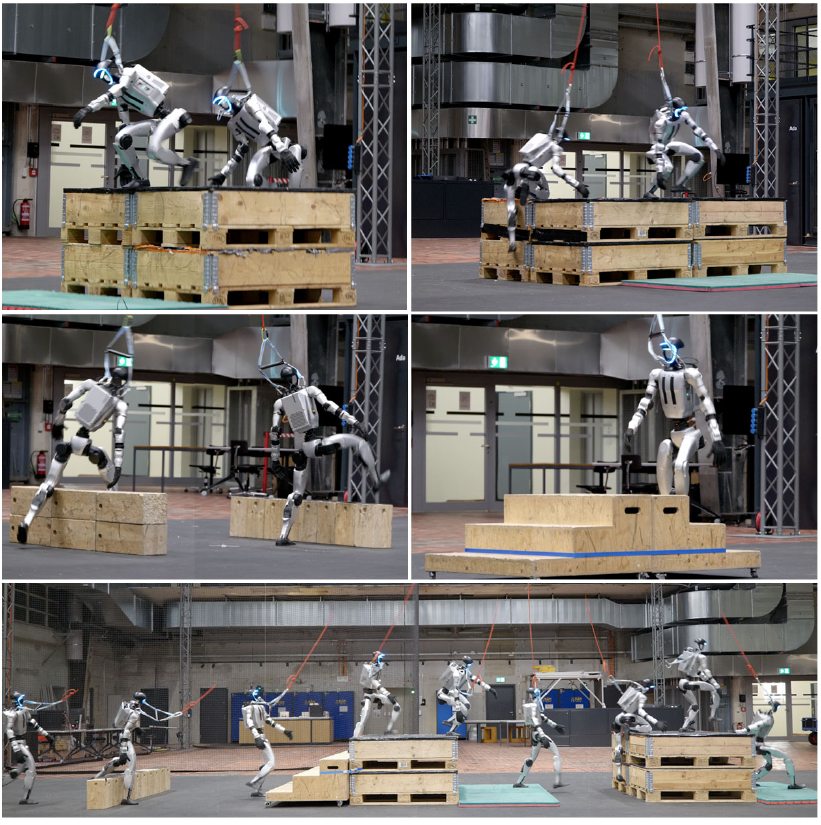}
    \caption{Whole-body humanoid locomotion on the Unitree G1 robot enabled by diffusion-based motion generation and RL-based motion tracking. Our hardware experiments show behaviors including box climbing, box descent, vaulting, and traversal over mixed terrain sequences. \url{https://wholebodylocomotion.github.io/}.
    }
    \vspace{-0.4cm}
    \label{fig:cover}
\end{figure}

Alternatively, generative models have recently shown strong potential for scalable motion control. 
Training generative models on large-scale motion datasets has proven effective in both character control~\citep{diffusecloc, mdm, closd} and robot control~\citep{diffusionpolicy, beyondmimic}, naturally reducing the engineering effort required as the motion repertoire expands. 
However, directly deploying expressive generated kinematic trajectories on real humanoid locomotion systems may introduce motion artifacts, such as foot sliding or temporal discontinuities, which can lead to catastrophic failure. 
To mitigate such issues, recent work~\citep{closd, robotmdm, parc} combines diffusion-based motion generation with RL-based motion tracking, allowing the tracker to output physically plausible actions based on kinematic generation. 
Despite this progress, their applicability to real humanoid robots in whole-body locomotion tasks remains insufficiently explored, and their deployment may potentially be hindered by slow inference and limited robustness to real-world disturbances.

To address these limitations, we propose a whole-body humanoid locomotion system that adapts a motion generation and motion tracking framework to real-time robot control for challenging humanoid locomotion tasks. 
In particular, we repurpose coordinated human whole-body skills learned from motion data for perceptive humanoid locomotion, and build a control framework that combines diffusion-based motion generation with RL-based motion tracking. 
The resulting system supports directional goal-reaching control together with terrain-aware whole-body adaptation. 
Our hardware experiments show that our approach enables efficient skill execution and generalization across diverse terrains, including terrains unseen during training.
Our results further demonstrate the effectiveness of using a motion generator trained on human motion data to guide an RL-based controller through challenging terrains requiring diverse skills, without heavily engineered distillation strategies.

\section{Related work} \label{sec:related_work}
\subsection{Whole-body Motion Tracking for Humanoids} 
In simulated whole-body character control, prior works \citep{deepMimic, 10.1145/3274247.3274506} have explored RL-based frameworks in which characters learn to imitate reference motion clips.
By training a control policy to minimize the discrepancy between the reference motion and the executed motion, the controller is able to reproduce a wide range of dynamic maneuvers while remaining consistent with the physical constraints of the controlled system.

More recently, this paradigm has been widely adopted in quadruped and humanoid robot control~\citep{omniretarget, beyondmimic, luo2025sonic, wasabi}. 
By designing imitation rewards and regularization terms, and optimizing the policy with the PPO algorithm~\citep{PPO}, recent works have demonstrated strong capability in imitating large-scale motion datasets and even enabling real-time teleoperation~\citep{twist, twist2, H2O}. 
However, for general locomotion tasks, pure motion tracking remains fundamentally limited in its adaptability to varying terrains.

\subsection{Motion Generation for Motion Control}\label{sec:motion_gen}
Motion generation with generative models provides an effective way to sample feasible motion trajectories from curated motion datasets.
Prior work~\citep{mdm} has demonstrated that diffusion models can generate smooth kinematic motion trajectories conditioned on text descriptions by incorporating additional geometric losses.
Building on this line of work, Tevet et al.~\citep{closd} further extend it to physical character control by coupling the generated motion with an RL-based motion tracker, which helps correct artifacts introduced by the generative model.
Serifi et al.~\citep{robotmdm} also combine motion generation with RL-based motion tracking, and further fine-tune the generator using value information from the critic network of the tracking policy.
Xu et al.~\citep{parc} adopt a similar design philosophy to \citep{closd} and extend it to humanoid parkour in simulation, demonstrating impressive whole-body motion control conditioned on surrounding terrain observations.
However, these methods mainly target simulated character control, and may still be constrained by the inference latency of diffusion models and lack of robustness for real robotic systems.
As a result, relatively few real-time whole-body humanoid locomotion systems with integrated exteroceptive sensing have been demonstrated.

Another line of work~\citep{diffusecloc, beyondmimic} generates joint state-action pairs without relying on a separate motion tracker.
These methods have been demonstrated on humanoid robots for tasks such as waypoint navigation and motion inpainting.
However, because control is produced directly from the co-diffusion process, they may still suffer from motion artifacts that can lead to unstable locomotion.
Moreover, more dynamic and demanding tasks, such as humanoid whole-body locomotion, are not well-studied in this setting. 
In our work, we follow the general framework of \citep{parc}, and adapt it to real-world humanoid locomotion tasks that require both whole-body coordination and real-time reaction to terrain conditions. 

\subsection{Skill Composition for Perceptive Locomotion Control}
Composing multiple skills into a unified policy is often regarded as an effective shortcut toward whole-body visuomotor locomotion control, compared with training a single policy entirely from scratch.
Hoeller et al.~\citep{ANYmalParkour} first train multiple expert skills for quadrupedal locomotion across different terrain types, and then learn a high-level planner that switches categorically among the expert policies. 
In contrast, Rudin et al.~\citep{parkourinthewild} employ DAgger~\citep{dagger} to first distill several expert policies into a single policy, and subsequently fine-tune the distilled policy with RL on new terrains.

For humanoid whole-body locomotion, concurrent work~\citep{APEX} adopts a similar strategy to~\citep{parkourinthewild}: it first trains terrain-specific expert policies from scratch with RL and then distills them into a unified student policy.
Wu et al.~\citep{PHP} likewise apply a similar distillation framework, but construct the expert skills from imitation-based RL policies trained with whole-body motion references.
These methods have shown promising results for terrain traversal requiring coordinated whole-body control.
However, both approaches may rely on carefully designed distillation procedures, such as expert assignment and data distribution design, to efficiently learn transitions and combinations across different skills.

In our work, instead of explicitly engineering a distillation pipeline over multiple expert skills, we use a diffusion-based motion generator trained on retargeted human motion data as a skill composition module that produces corresponding motion references for a low-level tracking policy based on the environment conditions.
In addition, to reduce the impact of motion artifacts introduced by the generator and improve control performance, we further fine-tune the motion tracking policy with RL under randomized direction commands and more diverse terrain configurations while using a frozen motion generator pretrained on the offline motion dataset.
More details are provided in \secref{sec:finetuning}.

\section{Method} \label{sec:method}
\begin{figure*}
    \centering
    \includegraphics[width=1\linewidth]{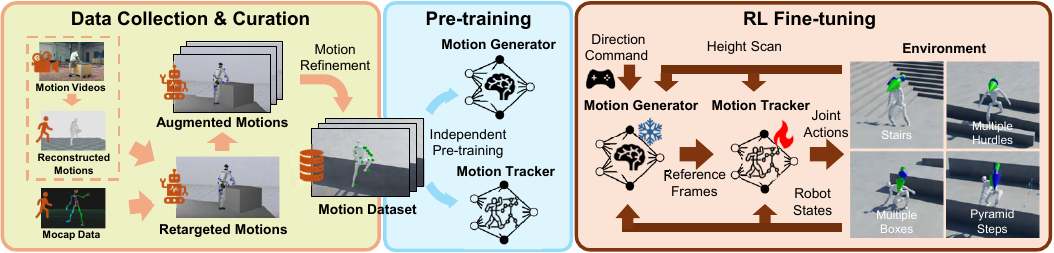}
    \caption{Overview of the training framework. (a) Data Collection \& Curation: whole-body robot motions are obtained from human motions through motion reconstruction and retargeting, followed by motion augmentation. (b) Pre-training: the diffusion-based motion generator and RL-based motion tracker are trained solely on the \textbf{offline motion dataset} from stage (a). (c) RL Fine-tuning: the motion tracker is \textbf{fine-tuned with RL} under randomized direction commands and on more diverse terrains with randomized configurations, including stairs, multiple vaulting hurdles, boxes, and pyramid-shaped high steps, while running the frozen motion generator.}
    \vspace{-0.3cm}
    \label{fig:framework}
\end{figure*}

The overall framework is illustrated in \figref{fig:framework}. 
Our method consists of three stages: 
(a) collecting whole-body motion data for different terrains and converting them into dynamically feasible robot trajectories;
(b) pre-training the RL-based motion tracker and the diffusion-based motion generator on this dataset; 
and (c) fine-tuning the motion tracker on more diverse terrain configurations, while keeping the motion generator frozen and running it in a receding-horizon manner during deployment.
We describe each stage below.

\subsection{Data Collection \& Curation}\label{sec:data_collection}
Our initial motion data consist of approximately 5 minutes of motion clips collected from two sources: motion videos captured by ourselves and public motion datasets~\cite{AMASS,omniretarget}. 
The dataset contains one representative motion type for each terrain skill, namely climbing onto a 50\,cm box, vaulting over a 35\,cm hurdle, jumping down from a 50\,cm box, and stair ascent and descent on steps of approximately 20\,cm height, together with omnidirectional walking motions such as straight walking and turning.
To enlarge this limited initial dataset and improve terrain diversity, we further apply motion augmentation.
In total, the resulting dataset contains approximately one hour of motion data for pre-training the motion generator and tracker.

\subsubsection{Initial Dataset Construction}
For motion skills recorded from videos, we first reconstruct human motions from raw videos using GVHMR~\citep{GVHMR}. We then retarget the reconstructed motions both from videos and motion datasets to the humanoid robot using a contact-constrained IK solver~\cite{drake}, similar to~\cite{omniretarget}.
After obtaining the retargeted robot motions, we manually place terrain objects to align with the original motions and preserve the physical interaction between the robot and the environment. 
To ensure that the final training data are physically plausible for both the motion tracker and the motion generator, we do not directly use the raw retargeted trajectories; instead, we refine training motions from a DeepMimic-style tracking policy trained on those retargeted motions.

\subsubsection{Motion Augmentation}
To further improve the generalization of both the motion tracker and the motion generator, we enlarge the dataset through kinematics-based augmentation. 
Specifically, we augment existing motions by varying terrain geometry, including scaling obstacle heights and inserting random small boxes along the motion path, and then optimizing the motions to remain consistent with the modified terrain. 
The optimization includes losses such as terrain penetration and motion smoothness losses to reduce collisions and discontinuities~\citep{parc}. 
As in the initial dataset construction, we do not use the optimized trajectories directly; instead, we train a tracking policy on them and record the resulting physically feasible trajectories. 
As a result, the augmented dataset covers box climbing and jump-down motions on 35--75\,cm boxes, vaulting motions over 25--45\,cm hurdles, stair motions on steps of 15--20\,cm height, and omnidirectional walking motions with randomly placed small boxes.

\subsection{Pre-training Stage}
The pre-training stage equips both the motion generator and the motion tracker with the basic ability to react to different environments and to output appropriate motion references and joint actions.

\subsubsection{Whole-body Motion Tracker}
We first train a single whole-body motion tracker on a single RTX 4090 GPU using a DeepMimic-style RL framework with PPO in IsaacLab~\citep{isaaclab}, based on the motion data collected in \secref{sec:data_collection}.
We design imitation rewards $r_{\rm{mimic}}$ and regularization terms $r_{\rm{reg}}$ (see \tabref{tab:reward_tracker}) so that the policy can imitate all reference motions while respecting the physical constraints of the real hardware:
\begin{equation}
    R_{\rm{pre}}=r_{\rm{mimic}} + r_{\rm{reg}}.
    \label{eq:reward_pretraining}
\end{equation}

The tracker observation consists of the reference state, including linear and angular velocity in $\mathbb{R}^3$, joint position and velocity in $\mathbb{R}^{23}$, and key body positions in the base frame in $\mathbb{R}^{9\times3}$; proprioceptive terms, including five consecutive frames of base angular velocity, projected gravity in $\mathbb{R}^3$, joint position and velocity, and the previous action in $\mathbb{R}^{23}$; together with terrain height scans in $\mathbb{R}^{17\times11}$. 
The policy outputs 23-dimensional target joint positions as actions.
Although Yang et al.~\cite{omniretarget} show that terrain information is not strictly required when the demonstrations are sufficiently accurate, we find that it can be beneficial to include exteroceptive input already at this stage, as it helps the later fine-tuning stage.

\subsubsection{Diffusion-based Motion Generator}
Our motion generator is based on a diffusion model that predicts whole-body reference motion sequences, following the architecture proposed in MDM~\citep{mdm, parc}.
The model predicts future motion features over a 0.5-second horizon (25 frames), including root position in $\mathbb{R}^3$, root orientation in $\mathbb{R}^4$, joint positions in $\mathbb{R}^{23}$, and body link positions in $\mathbb{R}^{23 \times 3}$, conditioned on the target heading vector, terrain height scan, and the past two frames of motion features.

During training, we randomly sample motion sequences from the dataset, compute the heading vector from the base pose difference, and use the first two frames as conditions to predict the remaining frames. The reconstruction loss is then computed by comparing the predicted motion with the ground-truth sequence.
In addition to the reconstruction loss, we incorporate geometric losses including velocity, joint consistency, and terrain penetration losses, similar to~\citep{parc}.
The generator is trained on the collected motion dataset and later serves as the motion planner that provides reference motions for the tracking policy over different terrains. 
To improve robustness against noisy robot states, we additionally perturb the height scans and previous-state conditions during training.

However, as discussed in \secref{sec:motion_gen}, generated motions may still contain artifacts, which can lead to failure when directly paired with a tracker trained mainly on high-quality references. 
Simply combining the pretrained motion generator and motion tracker can therefore result in limited robustness and reliability.

\subsection{RL Fine-tuning Stage} \label{sec:finetuning}
In this stage, we improve the robustness and reliability of the pretrained motion generator and motion tracker, and adapt the full control system for real-time closed-loop deployment on a real humanoid robot. 
Directly combining the pretrained generator and tracker may still suffer from motion artifacts, tracking failures caused by imperfect references, and limited generalization to terrains and target directions beyond those covered by the training dataset. 
Therefore, we introduce an additional fine-tuning stage in which the motion tracker is fine-tuned with the frozen motion generator via RL on more diverse terrains and under varied target direction conditions.

During this stage, the motion generator produces reference frames conditioned on the past two frames of robot states to form a closed-loop motion prediction process, rather than relying on autoregressive feedback of its own previous predictions. 
To reduce deployment latency under limited onboard computation, we use only 2 denoising steps in the motion generator.
We also inject additional noise into both the motion generator inference process and the tracker observations to improve robustness against real-world disturbances. 

In addition, we randomize the target heading direction and introduce a heading tracking reward as a task reward $r_{\rm{task}}$ (see \tabref{tab:reward_tracker}), together with the imitation $r_{\rm{mimic}}$ and regularization terms $r_{\rm{reg}}$, to encourage the full system to follow the desired direction while leveraging the motion priors acquired during pre-training:
\begin{equation}
    R_{\rm{post}} = r_{\rm{mimic}} + r_{\rm{reg}} + r_{\rm{task}}.
    \label{eq:reward_post}
\end{equation}
We further introduce more diverse training terrains, including stairs with 15--25\,cm step heights, continuous vaulting hurdles of 25--55\,cm height, and multiple climbing boxes and pyramid steps ranging from 30--85\,cm in height with varying widths as well as yaw and pitch angles.

After fine-tuning, the tracking policy effectively acts as a motion filter: it tracks the reference produced by the generator, while adjusting its execution to suppress unsafe behaviors using exteroceptive terrain observations.
Despite the reduced number of denoising steps, which may further degrade raw generation quality, the fine-tuned tracker still robustly tracks
the motion patterns produced by the generator and successfully traverses diverse terrains.
Randomization of target headings and terrain configurations further enables behaviors beyond those present in the offline data, such as corner box climbing, reorientation on the box before descent, and continuous vaulting.

\subsection{Details for Hardware Deployment}
For hardware experiments, the entire pipeline is deployed fully onboard the Unitree G1 robot. 
We use DLIO~\citep{chen2022dlio} with the onboard Livox MID360 LiDAR and IMU to estimate the robot base pose, which is used as an input to motion generation. 
For terrain perception, we employ Elevation Mapping CuPy~\citep{elevationmapping} together with the DLIO base pose to reconstruct terrain height information over different real-world environments.

Since the neck joint of the G1 robot is passive, aggressive motions can induce head movement that degrades base pose estimation. 
To improve state-estimation stability, we fuse the LiDAR IMU and torso IMU measurements by subtracting the orientation estimated by DLIO from the moving-average pitch measured by the torso IMU, which allows us to estimate the head pitch variation and compensate for its effect.

To accelerate diffusion-based motion generation, we deploy the generator with TensorRT, reducing inference time to approximately 0.02\,s. 
Although the motion generator operates with a 0.5-second prediction horizon (2 Hz) during fine-tuning, we update the reference every 0.25 seconds in a receding-horizon way during deployment to improve responsiveness to the surrounding environment.
The motion generator runs on a Jetson Thor mounted on the back of the robot, while the motion tracking controller and the remaining onboard modules run on the Jetson Orin integrated into the Unitree G1 platform.

\begin{table}[h!]
    \centering
    \scriptsize
    \setlength{\tabcolsep}{2pt}
    \renewcommand{\arraystretch}{1.15}
    \setlength{\extrarowheight}{1pt} 
    \caption{Reward Equations for RL-based Motion Tracker}
    \vspace{-0.2cm}
    \label{tab:reward_tracker}
    \begin{tabular}{p{0.08\columnwidth}|p{0.31\columnwidth}|p{0.56\columnwidth}}
    \hline
    Group & Name & Equation \\ \hline

    \multirow{6}{*}{$r_{\mathrm{mimic}}$}
    & Base Pose Tracking
    & $3\exp\!\left(-4\norm{\mathbf{p}_{b}^w-\mathbf{p}_{b}^{w*}}^2
       - \norm{\mathbf{q}_b \otimes {\mathbf{q}_b^{*}}^{-1}}^2\right)$ \\

    & Base Velocity Tracking
    & $\exp\!\left(-\norm{\mathbf{v}_b^b-\mathbf{v}_b^{b*}}^2
       - 0.1\norm{\boldsymbol{\omega}_b^b-\boldsymbol{\omega}_b^{b*}}^2\right)$ \\

    & Joint Position Tracking
    & $2.5\exp\!\left(-2\sum_{i=1}^{23}0.25(q_{i}-q_{i}^{*})^2\right)$ \\

    & Joint Velocity Tracking
    & $0.5\exp\!\left(-2\sum_{i=1}^{23}0.01(\dot{q}_{i}-\dot{q}_{i}^{*})^2\right)$ \\

    & Body Pos Tracking (Base)
    & $2\exp\!\left(-\sum_{i=1}^9\norm{\mathbf{p}_{l,i}^b-\mathbf{p}_{l,i}^{b*}}^2\right)$ \\

    & Body Pos Tracking (Global)
    & $\exp\!\left(-\sum_{i=1}^9\norm{\mathbf{p}_{l,i}^w-\mathbf{p}_{l,i}^{w*}}^2\right)$ \\

    \hline

    \multirow{7}{*}{$r_{\mathrm{reg}}$}
    & Action Rate (First-Order)
    & $-0.1\sum_{i=1}^{23}(a_{t,i}-a_{t-1,i})^2$ \\

    & Action Rate (Second-Order)
    & $-0.05\sum_{i=1}^{23}(a_{t,i}-2a_{t-1,i}+a_{t-2,i})^2$ \\

    & Joint Position Limits
    & $-10\sum_{i=1}^{23}\max(q_{i}-q_{\mathrm{lim},i},0)$ \\

    & Joint Velocity Limits
    & $-\sum_{i=1}^{23}\max(\dot{q}_{i}-\dot{q}_{\mathrm{lim},i},0)$ \\

    & Torque Limits
    & $-0.001\sum_{i=1}^{23}\max(\tau_{i}-\tau_{\mathrm{lim},i},0)$ \\

    & Joint Torques
    & $-0.00001\sum_{i=1}^{23}\tau_{i}^2$ \\

    & Body Linear Acceleration
    & $-0.01\sum_{i=1}^9\norm{\ddot{\mathbf{p}}_{l,i}^{w}}^2$ \\

    \hline
    $r_{\mathrm{task}}$ & Target Heading Tracking
    & $1.5\langle \mathbf{v}_b^b, \mathbf{d}_b \rangle / \norm{\mathbf{v}_b^b}$ \\
    \hline
    \end{tabular}
    \vspace{-0.4cm}
\end{table}
\begin{table}[ht!]
    \centering
    \caption{Symbols used in \tabref{tab:reward_tracker}}
    \vspace{-0.2cm}
    \label{tab:symbol}
    \begin{tabular}{l|l}
    \hline
    Symbol & Description \\ \hline
    $(\cdot)^*$ & Reference quantity \\
    $(\cdot)^b$ & Expressed in the base frame \\
    $(\cdot)^w$ & Expressed in the world frame \\
    $\mathbf{p}_b$ & Base position \\
    $\mathbf{q}_b$ & Base orientation quaternion \\
    $\mathbf{v}_b$ & Base linear velocity \\
    $\boldsymbol{\omega}_b$ & Base angular velocity \\
    $\mathbf{d}_b$ & Base target heading \\
    $\mathbf{p}_{l,i}$ & Position of key body $i$ \\
    $\ddot{\mathbf{p}}_{l,i}$ & Acceleration of key body $i$ \\
    $q_i$ & Position of joint $i$ \\
    $\dot{q}_i$ & Velocity of joint $i$ \\
    $\tau_i$ & Torque of joint $i$ \\
    $q_{\mathrm{lim},i}$ & Joint position limit of joint $i$\\
    $\dot{q}_{\mathrm{lim},i}$ & Joint velocity limit of joint $i$\\
    $\tau_{\mathrm{lim},i}$ & Joint torque limit of joint $i$\\
    \hline
    \end{tabular}
    \vspace{-0.5cm}
\end{table}

\section{Experiments} \label{sec:results}
\begin{figure*}
    \centering
    \includegraphics[width=1\linewidth]{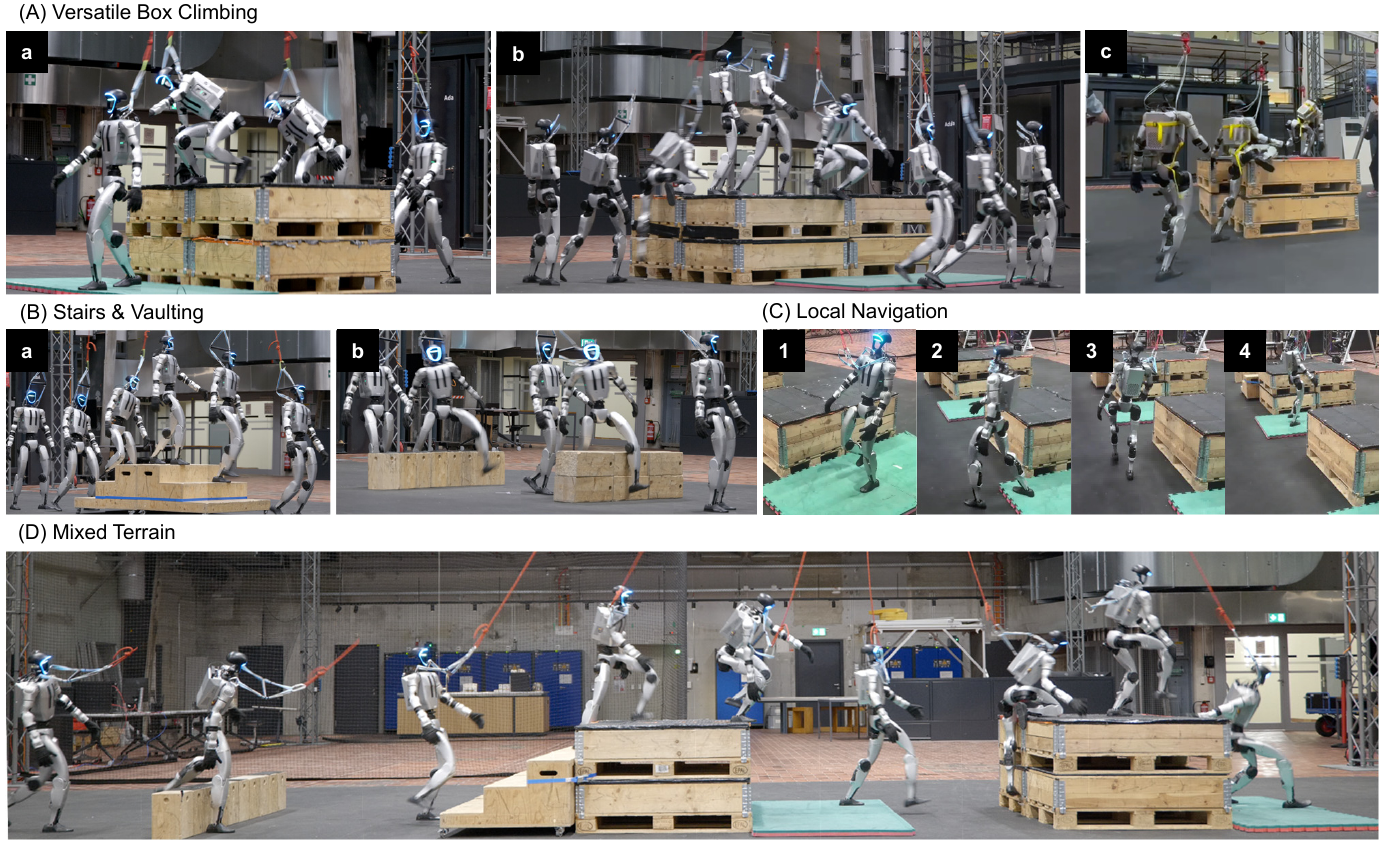}
    \caption{Results of hardware experiments.
    (A) The robot climbs onto a 75\,cm box and jumps down in three different ways: (a) straight ascent and descent; (b) straight ascent followed by a $90^\circ$ right turn and side jump-down; and (c) ascent and descent from the box corner.
    (B) The robot traverses (a) a staircase and (b) a sequence of vaulting hurdles. 
    (C) The robot shows local navigation behavior by bypassing the box to reach the target position.
    (D) The robot traverses a mixed terrain sequence consisting of a vaulting obstacle, stairs, and box climbing.}
    \label{fig:hardware_results}
    \vspace{-0.3cm}
\end{figure*}

In this section, we present the experimental results of the proposed humanoid locomotion system in simulation and on real hardware. 
We also provide quantitative evaluation to examine the contribution of motion generation to generalization, and the effect of RL fine-tuning of the motion tracker on robustness and success rates across terrain types.

\subsection{Hardware Results}
We evaluate the proposed pipeline on real hardware as shown in \figref{fig:cover}, \ref{fig:hardware_results}, and the supplementary video. 
The hardware experiments demonstrate both the effectiveness of our controller and its generalization ability across diverse terrains.
In particular, we evaluate the controller on a range of terrain types, including box ascent and descent, stair ascent and descent, and continuous vaulting over obstacles of varying heights, as well as on more challenging compound terrain settings that combine vaulting, stairs, and box climbing within a single run.
For box ascent and descent, our controller exhibits versatile climbing behaviors, including box traversal from different approach directions as well as the ability to traverse a box with on-top reorientation before descent. 
During ascent, the robot uses its knees and hands for support, while during descent it jumps off the box and uses the hands for cushioning, consistent with the motion style observed in the training data.
For vaulting, the robot can traverse multiple hurdles of different heights, typically crossing over them directly rather than stepping onto the hurdle first. 
For stairs, the controller switches to stair ascent and descent behaviors. 
When these terrain types are combined, the controller can also transition dynamically among different motion styles according to the terrain, including sequences such as stair ascent following a jump-down motion, even though such terrain combinations are not seen during RL fine-tuning.

In addition, as shown in \figref{fig:hardware_results}C and the supplementary video, we observe a form of local navigation behavior in the system after RL fine-tuning: when the robot is not in a suitable state to traverse an obstacle under the direction command computed from the given goal position using the motion generated by the planner, the tracking policy may partially override the reference motion and instead bypass the obstacle from the side to avoid failure and still reach the goal.

\subsection{Quantitative Results}
\begin{figure*}[t!]
    \centering
    \includegraphics[width=1\linewidth]{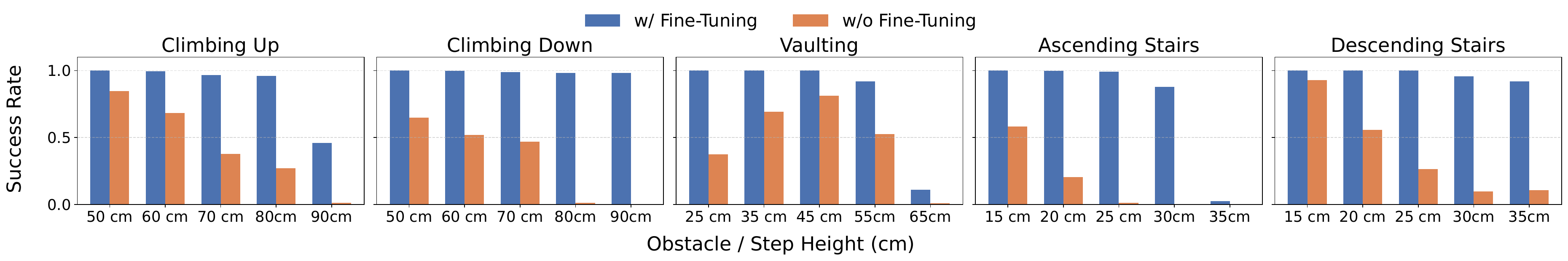}
    \caption{Comparative results of the success rates of the motion tracker with and without fine-tuning across five terrain traversal tasks, including climbing up/down, vaulting, and stair ascent/descent. 
    For each task, we spawn 500 robots in simulation with randomized initial poses and assign each of them a goal position that guides traversal of the corresponding terrain obstacle, from which the target direction is computed. 
    The results consistently show that fine-tuning the tracker with a frozen motion generator improves the success rate of reaching the goal.}
    \vspace{-0.4cm}
    \label{fig:comparison_finetuning}
\end{figure*}

We further analyze two key design choices of our framework through quantitative simulation results: whether online motion generation improves adaptation and generalization across diverse terrains at deployment, and whether the additional fine-tuning stage is necessary after coupling the generator with the motion tracker. 
As shown in \tabref{tab:comparison_motion_gen} and \figref{fig:comparison_finetuning}, both components are important for the final system.

\subsubsection{Benefit of Online Motion Generation}

\begin{table*}[t!]
\centering
\caption{Comparative Success Rates of Fixed-Reference Motion Tracking and The Proposed Full System Across Three Evaluation Settings.}
\label{tab:comparison_motion_gen}
\vspace{-0.2cm}
\subfloat[Box Climbing \label{tab:box_climb}]{
\resizebox{0.265\textwidth}{!}{
\begin{tabular}{c|c|c|c}
\hline
Box Height & Yaw & \multirow{2}{*}{Tracker Only} & \multirow{2}{*}{Tracker+Gen.} \\
$[\mathrm{cm}]$ & $[\mathrm{deg}]$ & & \\
\hline
40 & 0   & 0.982 & 1.000 \\
\hline
\multirow{5}{*}{50} & -30 & 0.988 & 0.980 \\
\cline{2-4}
 & 30  & 1.000 & 0.982 \\
\cline{2-4}
 & -15 & 0.974 & 1.000 \\
\cline{2-4}
 & 15  & 0.970 & 0.998 \\
\cline{2-4}
 & 0   & 0.998 & 1.000 \\
\hline
60 & 0 & 0.984 & 0.996 \\
\hline
70 & 0 & 0.606 & 0.966 \\
\hline
80 & 0 & 0.230 & 0.962 \\
\hline
\multicolumn{2}{c|}{Mean $\pm$ SD} & 0.859 $\pm$ 0.252 & \textbf{0.987 $\pm$ 0.014} \\
\hline
\end{tabular}
}}
\hspace{0.01\textwidth}
\subfloat[Vaulting \label{tab:vault}]{
\resizebox{0.3\textwidth}{!}{
\begin{tabular}{c|c|c|c}
\hline
Hurdle Height & Yaw & \multirow{2}{*}{Tracker Only} & \multirow{2}{*}{Tracker+Gen.} \\
$[\mathrm{cm}]$ & $[\mathrm{deg}]$ & & \\
\hline
25 & 0   & 0.978 & 1.000 \\
\hline
\multirow{5}{*}{35} & -40 & 0.808 & 1.000 \\
\cline{2-4}
 & 40  & 0.492 & 1.000 \\
\cline{2-4}
 & -20  & 0.922 & 1.000 \\
\cline{2-4}
 & 20  & 0.954 & 1.000 \\
\cline{2-4}
 & 0   & 0.988 & 1.000 \\
\hline
45 & 0   & 0.950 & 1.000 \\
\hline
55 & 0   & 0.348 & 0.920 \\
\hline
\multicolumn{2}{c|}{Mean $\pm$ SD} & 0.805 $\pm$ 0.231 & \textbf{0.990 $\pm$ 0.026} \\
\hline
\end{tabular}
}}
\hspace{0.01\textwidth}
\subfloat[Ascending Stairs \label{tab:stairs_ascend}]{
\resizebox{0.32\textwidth}{!}{
\begin{tabular}{c|c|c|c}
\hline
Step Height & Yaw & \multirow{2}{*}{Tracker Only} & \multirow{2}{*}{Tracker+Gen.} \\
$[\mathrm{cm}]$ & $[\mathrm{deg}]$ & & \\
\hline
15 & 0   & 1.000 & 1.000 \\
\hline
\multirow{5}{*}{20} & -30 & 0.948 & 1.000 \\
\cline{2-4}
 & 30  & 0.916 & 0.994 \\
\cline{2-4}
 & -15 & 0.970 & 1.000 \\
\cline{2-4}
 & 15  & 0.990 & 0.996 \\
\cline{2-4}
 & 0   & 0.980 & 1.000 \\
\hline
25 & 0   & 0.114 & 0.986 \\
\hline
\multicolumn{2}{c|}{Mean $\pm$ SD} & 0.845 $\pm$ 0.300 & \textbf{0.997 $\pm$ 0.005} \\
\hline
\end{tabular}
}}
\end{table*}

We first evaluate whether online motion generation is beneficial for enabling the final system to adapt to diverse terrain variations by comparing it with fixed-reference tracking.
We compare two settings on three representative tasks: box climbing, vaulting, and stair ascent. 
For box climbing, the reference trajectory is a motion for climbing onto a 50\,cm box; for vaulting, the reference is a trajectory for traversing a 35\,cm-high and 20\,cm-wide hurdle; for stair ascent, the reference is a trajectory for ascending a staircase with 30\,cm-wide and 20\,cm-high steps. 
We then modify the terrain geometry at test time by changing the obstacle height or rotating the terrain mesh in yaw. 
\textbf{Tracker Only} uses the fine-tuned motion tracker with a fixed reference trajectory, and success is measured by whether the robot reaches the final XY base position of the reference trajectory. 
\textbf{Tracker + Gen.} uses the proposed full system, where the motion generator produces reference motions online from the target direction and terrain input, and success is measured by whether the robot reaches the goal position placed on the terrain. 
The goal position is chosen to match the final base position of the reference trajectory used in the \textit{Tracker Only} setting, and the direction command is computed online accordingly. 
For each task, 500 robots are spawned in simulation with randomized initial poses and joint positions.

As shown in \tabref{tab:comparison_motion_gen}, the full system with online motion generation achieves better generalization and higher average success rates across all three tasks than fixed-reference tracking on the corresponding terrain settings.
For box climbing, the \textit{Tracker Only} baseline performs well close to the nominal reference condition, but becomes brittle as the obstacle height increases.
In contrast, \textit{Tracker + Gen.} remains consistently robust across all tested settings. 
For vaulting, the fine-tuned tracker already exhibits some local generalization and can handle different hurdle heights, but its performance still degrades under larger changes in terrain height and orientation.
The full system, by comparison, remains stable across all evaluated cases. 
A similar pattern appears in stair ascent: while the fixed-reference tracker can tolerate yaw perturbations, it deteriorates markedly under the height increase, whereas the full system continues to achieve nearly perfect performance.

These results highlight two points. 
First, the fine-tuned tracker is not entirely rigid. 
Because it observes terrain height scans and is fine-tuned on diverse terrain configurations, it can retain a limited degree of local terrain generalization. 
This is particularly visible in vaulting and box climbing, where nearby obstacle scales can sometimes be handled even without adapting the reference motion. 
Second, this limited generalization is fundamentally insufficient for robust perceptive locomotion. 
Once the terrain geometry changes more substantially, especially in ways that require changing the timing or style of the motion rather than merely tolerating small mismatch, the fixed-reference tracker degrades rapidly. 
During deployment, terrain-aware reference motions produced online by the motion generator allow the full system to better adapt its traversal behavior to diverse obstacle geometries.
Overall, the results in \tabref{tab:comparison_motion_gen} highlight the importance of online motion generation for adaptation and generalization across diverse terrains in the final system.

\subsubsection{Impact of Fine-Tuning the Motion Tracker}
The above results show that online motion generation is important for generalization during deployment. 
We next ask whether the additional fine-tuning stage is also necessary, or whether the pretrained tracker can simply be combined with the motion generator directly. 
To answer this question, we compare the pretrained tracker and the fine-tuned tracker under the same motion generator, as shown in \figref{fig:comparison_finetuning}. 
We evaluate five terrain traversal tasks, namely climbing up, climbing down, vaulting, ascending stairs, and descending stairs. 
For each task, a goal position is placed behind the terrain obstacle, from which the direction command is computed, and 500 robots are spawned in simulation to measure the success rate.

\Figref{fig:comparison_finetuning} shows that fine-tuning the tracker consistently improves success rates across all evaluated tasks, with the largest gains appearing on the more difficult terrain settings. 
For climbing up and climbing down, the improvement becomes more apparent as the box height increases, indicating that fine-tuning is especially important when the robot is required to react robustly to more demanding vertical transitions. 
For stair ascent and descent, the fine-tuned tracker also shows clear advantages on larger step heights.
Vaulting is comparatively easier than the other tasks, but the fine-tuned tracker still remains more reliable overall. 
Taken together, the histogram suggests that fine-tuning the tracker with RL does not merely provide marginal refinement, but systematically improves robustness across a broad range of traversal tasks.

A likely reason is that the pretrained tracker is optimized on a fixed set of offline motion trajectories, which are relatively smooth and physically filtered. 
During deployment, however, the online references generated by the motion generator are conditioned on noisy robot state, target direction, and terrain observations, and may therefore contain temporal discontinuities or motion artifacts. 
This creates a distribution mismatch between the tracker training references and the actual reference stream encountered during control. 
Fine-tuning with the frozen generator is therefore crucial: it allows the tracker to adapt to the distribution and imperfections of generated references, learn how to make effective use of motion patterns for task completion, and suppress unsafe behaviors that could otherwise lead to failure. 
This observation is also consistent with the qualitative behavior discussed in \secref{sec:finetuning}, where the fine-tuned tracker effectively acts as a motion filter over the generated references. 
Overall, \figref{fig:comparison_finetuning} shows that the fine-tuning stage is an important part of making the combined generation-and-tracking framework robust in practice.

\section{Conclusion} 
\label{sec:conclusion}
Overall, our results demonstrate that combining motion generation with motion tracking is an effective way to realize perceptive whole-body humanoid locomotion. 
The proposed system enables efficient traversal over diverse terrain types and terrain combinations unseen during training, while the quantitative analysis highlights the importance of both online motion generation and tracker fine-tuning for adaptation and robustness.

One limitation of the current system is that it relies on a LiDAR-based elevation mapping pipeline to reconstruct the surrounding environment. 
When the mapping quality degrades due to sensing noises, the locomotion performance can deteriorate substantially. 
A promising direction for future work is to adopt perception architectures that do not depend as strongly on high-quality geometric sensing, such as neural mapping~\cite{AME2}, or to incorporate a belief encoder~\cite{wildanymal} that compensates for sensing failures through proprioceptive observations. 
Another important direction is to extend the framework beyond pure locomotion to a broader range of loco-manipulation tasks and to more challenging outdoor environments.

\section*{Acknowledgments}
This work was supported in part by NCCR Automation, the NSERC Discovery Grant (RGPIN-2015-04843), RobotX AHL project, and ETH AI Center. The authors gratefully acknowledge Changan Chen, Ryan Batke, Yihan Wang, and Zhihao Cao for their valuable assistance with this project and the hardware experiments.

\bibliographystyle{IEEEtran}

\bibliography{references}

\end{document}